\pdfoutput=1

\documentclass[11pt]{article}

\usepackage[final]{acl}

\usepackage{times}
\usepackage{latexsym}

\usepackage[T1]{fontenc}

\usepackage[utf8]{inputenc}

\usepackage{microtype}

\usepackage{inconsolata}

\usepackage{graphicx}

\usepackage{algorithm}
\usepackage{algpseudocode}
\usepackage{amsmath}
\usepackage{amssymb}
\usepackage{booktabs}
\usepackage{CJK}
\usepackage{enumitem}
\usepackage{makecell}
\usepackage{multirow}
\usepackage{soul}
\usepackage{subcaption}
\usepackage{xurl}

\newcommand{\ours}{MixCal}

\newcommand{\result}[2]{#1\textsubscript{\,\textcolor{gray}{#2}}}
\newcommand{\referenceresult}[2]{\textcolor{gray}{#1}\textsubscript{\,\textcolor{gray}{#2}}}
\newcommand{\bestresult}[2]{\textbf{#1}\textsubscript{\,\textcolor{gray}{#2}}}

\newcommand{\chinese}[1]{\begin{CJK}{UTF8}{gbsn}#1\end{CJK}}

\title{Compressing Language Models for Specialized Domains}

\author{Miles Williams$^{\diamondsuit \spadesuit}$ \quad George Chrysostomou$^{\spadesuit}$ \quad Vitor Jeronymo$^{\spadesuit}$ \quad Nikolaos Aletras$^{\diamondsuit}$ \\ $^{\diamondsuit}$University of Sheffield\\$^{\spadesuit}$Enterprise AI Services, AstraZeneca\\ \texttt{\{mwilliams15, n.aletras\}@sheffield.ac.uk}}

\begin{document}
\maketitle

\begin{abstract}
Language models (LMs) excel at tasks across diverse domains, yet require substantial computational resources during inference. Compression techniques such as pruning and quantization offer a practical path towards efficient LM deployment, exemplified by their ability to preserve performance on general-purpose benchmarks. However, general-purpose LM compression methods can negatively affect performance in specialized domains (e.g. biomedical or legal). Recent work has sought to address this issue, but requires a computationally expensive full-parameter fine-tuning pipeline. To this end, we propose \ours{}, a novel calibration method designed to improve the in-domain performance of compressed LMs in a post-training setting. Through extensive experimentation, we demonstrate that \ours{} substantially outperforms existing approaches on domain-specific tasks and preserves general performance. Notably, these performance gains are achieved while also reducing the computational cost of LM compression.\footnote{\url{https://github.com/mlsw/domain-compression}}
\end{abstract}

\section{Introduction}

Language models (LMs) have demonstrated remarkable performance across tasks from a range of domains \citep{walsh-etal-2025-olmo, guo-etal-2025-deepseek, kamath-etal-2025-gemma3}. Behind this success lies a recipe with two key ingredients: highly parameterized models and extensive training. However, the vast scale of these models presents substantial challenges in their deployment and application \citep{treviso-etal-2023-efficient, zhu-etal-2024-survey-model}. \citet{luccioni-etal-2024-power} suggest that the trend towards \emph{general-purpose} models has introduced substantial yet potentially unnecessary inference costs.

Model compression techniques, such as quantization and pruning, are foundational approaches aimed at reducing the computational footprint of LMs during inference \citep{zhu-etal-2024-survey-model}. Quantization represents weights (and/or activations) with lower precision, while pruning removes less important weights. Notably, recent work has shown the successful application of quantization \citep{frantar-etal-2023-optq, lin-etal-2024-awq} and pruning \citep{frantar-alistarh-2023-sparsegpt, sun-etal-2024-simple} to general-purpose LMs without any additional training.

\begin{figure}
\centering
\includegraphics[width=\linewidth]{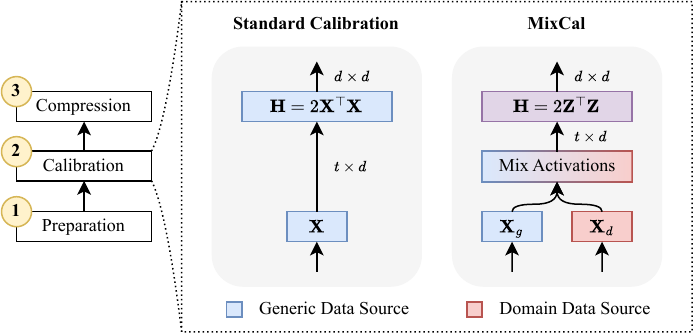}
\caption{\ours{} (\S \ref{sec:methodology}) is applied within the calibration phase of model compression, leveraging a combination of activations from generic and domain-specific data.}
\label{fig:diagram}
\end{figure}

LM compression studies typically focus on preserving general-purpose performance, i.e. language modeling and commonsense reasoning capabilities \citep{frantar-alistarh-2023-sparsegpt, ma-etal-2023-llm, sun-etal-2024-simple}. 
However, in practice, LMs may be deployed within only one particular domain, e.g. biomedical or legal \citep{labrak-etal-2024-biomistral, colombo-etal-2024-saullm, ling-etal-2024-domain, chen-etal-2024-survey}. This scenario unlocks new paths towards improving inference efficiency by extracting domain-specific LMs from general-purpose models.

\citet{zhang-etal-2024-pruning} proposed D-Pruner, a pruning method aiming to preserve weights that are influential to both domain-specific and general capabilities. 
To identify such weights, D-Pruner leverages the gradient information from a composite loss function that incorporates general weight importance scores. However, this requires a full-parameter fine-tuning pipeline for the LM, thus incurring substantial computational costs.

The majority of post-training LM compression methods rely upon \textit{calibration data}, a small amount of data used to aid the analysis of layer activations. Recent work has shown that calibration data can impact task performance \citep{williams-aletras-2024-impact, bandari-etal-2024-c4, williams-etal-2025-self}. Inspired by this line of work, we investigate how to effectively leverage domain-specific data for calibration, aiming to maximize in-domain performance without sacrificing general capabilities.
Our main contributions are as follows:

\begin{enumerate}[left=0pt,partopsep=0em]

\item We find that in-domain calibration data can play an important role in LM pruning, maximizing performance retention on domain-specific tasks.

\item We propose \ours{}, a novel calibration method for Hessian-based compression, enabling the effective use of in-domain calibration data (Figure~\ref{fig:diagram}). Our approach outperforms existing general and domain-specific pruning methods while offering greater computational efficiency.

\end{enumerate}

\section{Related Work}

\paragraph{Quantization.}

The objective of quantization is to represent weights (and optionally activations) using fewer bits. This reduction in precision reduces memory requirements, and typically enables inference speedups \citep{gholami-etal-2021-survey}. Beyond their scale, contemporary LMs pose unique challenges for effective quantization, including the existence of high-magnitude outlier features \citep{bondarenko-etal-2021-understanding, dettmers-etal-2022-gpt3}. Recent directions include: holding outlier weights in higher precision \citep{dettmers-etal-2022-gpt3}, Hessian-based weight sensitivity \citep{frantar-etal-2023-optq}, searching for optimal clipping thresholds \citep{wei-etal-2023-outlier}, or combinations of these approaches \citep{lin-etal-2024-awq, dettmers-etal-2024-spqr}.

\paragraph{Pruning.}

The aim of neural network pruning is to remove less important weights, therefore reducing the overall model size \citep{lecun-etal-1989-optimal}. Pruning can be performed at the level of individual weights (unstructured), within groups of weights (semi-structured), or entire dimensions (structured) \citep{han-etal-2015-learning, mishra-etal-2021-accelerating, ma-etal-2023-llm}. In particular, 2:4 semi-structured sparsity (i.e. pruning two weights in every block of four) enables enhanced inference performance on NVIDIA GPUs \citep{mishra-etal-2021-accelerating}. However, the extensive size of Transformer-based \citep{vaswani-etal-2017-attention} LMs presents challenges in pruning them optimally \citep{hassibi-etal-1993-optimal}. Recent work has instead decomposed LM pruning into a sequential layer-wise approach, demonstrating remarkable performance retention, even at high sparsity levels \citep{frantar-alistarh-2023-sparsegpt, sun-etal-2024-simple, yin-etal-2024-outlier}.

\paragraph{Domain-specific pruning.}

Early work focused on pruning deep neural networks for specific tasks \citep{han-etal-2015-learning, molchanov-etal-2017-pruning}. This trend continued \citep{sanh-etal-2020-movement, lagunas-etal-2021-block, kwon-etal-2022-fast} following the advent of BERT \citep{devlin-etal-2019-bert}. However, the shift towards general-purpose LMs \citep{brown-etal-2020-language, dredze-etal-2024-academics} has led to a focus on preserving general performance \citep{frantar-alistarh-2023-sparsegpt, ma-etal-2023-llm, sun-etal-2024-simple}, i.e. language modeling and reasoning. Most recently, \citet{zhang-etal-2024-pruning} proposed D-Pruner for domain-specific pruning, leveraging general weight importance to form a domain-specific training loss. However, this requires an expensive full-parameter fine-tuning pipeline, yet does not consistently outperform general-purpose pruning methods.

\paragraph{Calibration data.}

In a post-training setting, model compression usually relies upon calibration data \citep{wan-etal-2024-efficient}. Calibration data consists of a small number of unlabeled examples for the generation of layer activations \citep{nagel-etal-2020-up, hubara-etal-2021-accurate}. Typically, these examples are randomly sampled from web text or pre-training datasets (e.g. C4; \citealp{raffel-etal-2020-exploring}). Recent work has illustrated the influential role that calibration data can play, impacting the downstream performance of compressed models \citep{williams-aletras-2024-impact, williams-etal-2025-self}. However, \citet{bandari-etal-2024-c4} suggest that pruning using downstream task data does not necessarily outperform generic data on the corresponding task.

\section{\ours}
\label{sec:methodology}

\subsection{Preliminaries}

The Optimal Brain Surgeon (OBS; \citealp{hassibi-etal-1993-optimal}) algorithm leverages second-order derivatives to accurately prune weights from a neural network. These second-order derivatives, which indicate the curvature of the loss function with respect to the weights, are organized in a square matrix known as the Hessian $\mathbf{H}$. Based on the Hessian, the OBS algorithm iteratively removes the weight $w_m$ with the lowest saliency $\varepsilon_m$, followed by applying the optimal update for the remaining weights $\boldsymbol{\delta}_m$:
\begin{equation*}
\varepsilon_m = \frac{1}{2}\frac{w^2_m}{[\mathbf{H}^{-1}]_{mm}}, \quad \boldsymbol{\delta}_m = -\frac{w_m}{[\mathbf{H}^{-1}]_{mm}}\cdot\mathbf{H}^{-1}_{:,m}
\end{equation*}

\citet{frantar-alistarh-2022-optimal} reformulate pruning as a case of the layer-wise compression problem, while retaining the OBS weight update procedure. Given a layer with input activations $\mathbf{X}$, the aim is to minimize the error between the original layer weights $\mathbf{W}$ and newly compressed weights $\mathbf{\widehat{W}}$:
\begin{equation*}
\underset{\mathbf{\widehat{W}}}{\arg\min}\; \|\mathbf{X}\mathbf{W} - \mathbf{X}\mathbf{\widehat{W}}\|^2_F
\end{equation*}
As the layer input activations are derived from a fixed set of calibration data, the layer outputs (i.e. $\mathbf{Y}=\mathbf{XW}$) are also fixed. Consequently, the layer-wise Hessian is computed as:
\begin{equation*}
\mathbf{H} = 2\mathbf{X}^\top\mathbf{X}
\end{equation*}

\subsection{Motivation}

\begin{figure}[t]
\centering
\includegraphics{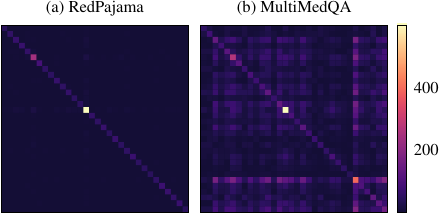}
\caption{The Hessian at layer 16 of Mistral NeMo 12B, computed with (a) generic calibration data, and (b) domain-specific calibration data. For clarity, we present the magnitude of the elements for the first 32 features.}
\label{fig:hessian-structure}
\end{figure}

Language models are trained over diverse and expansive corpora. Accordingly, a random sample of generic calibration data often proves sufficient to approximate the Hessian \citep{frantar-alistarh-2023-sparsegpt}. These generic samples can also help preserve general capabilities \citep{bandari-etal-2024-c4}. However, LM weights additionally encode domain-specific knowledge \citep{singhal-etal-2023-large}. For example, Figure~\ref{fig:hessian-structure} illustrates that a feature may be sensitive in a target domain yet appear unimportant under generic data. Therefore, we hypothesize that compressing LMs for specialized domains while maintaining general performance requires preserving weights that are important under both settings. To this end, we propose \emph{\ours{}} (Algorithm \ref{alg:ours}).

\subsection{Mixing Activations}

First, we extend layer-wise compression to a multi-objective setting. The standard layer-wise reconstruction loss for a layer with weights $\mathbf{W}$, compressed weights $\widehat{\mathbf{W}}$, and input activations $\mathbf{X}$ is:
\begin{equation*}
\mathcal{L}(\widehat{\mathbf{W}}, \mathbf{X}) = \| \mathbf{X}\mathbf{W} - \mathbf{X}\widehat{\mathbf{W}}\|^2_F
\end{equation*}
We introduce two distinct sources of calibration data, a domain-specific dataset $\mathcal{D}_d$, and a general-purpose dataset $\mathcal{D}_g$. Samples from these datasets are used to form the domain-specific and general-purpose activations, $\mathbf{X}_d$ and $\mathbf{X}_g$, respectively. We then define a combined loss, which balances reconstruction between both calibration datasets:
\begin{equation*}
\mathcal{L}_\text{weighted} = \alpha \mathcal{L}(\widehat{\mathbf{W}}, \mathbf{X}_d) + \beta \mathcal{L}(\widehat{\mathbf{W}}, \mathbf{X}_g)
\end{equation*}
To control the relative contribution from each loss term, we introduce the weight coefficients $\alpha$ and $\beta$. We impose $\alpha + \beta = 1$ to avoid an arbitrary scaling, setting $\beta = 1 - \alpha$. The loss is therefore parameterized by a single coefficient, $\alpha \in [0, 1]$. This loss function yields the following Hessian:
\begin{equation*}
\mathbf{H}_\text{weighted} = 2(\alpha\mathbf{X}_d^\top\mathbf{X}_d + \beta\mathbf{X}_g^\top\mathbf{X}_g )
\end{equation*}

However, this formulation treats the domain-specific and general-purpose activations independently. The Hessian $\mathbf{H}_\text{weighted}$ is a weighted sum of the domain-specific and general-purpose Hessians, namely $2\mathbf{X}_d^\top\mathbf{X}_d$ and $2\mathbf{X}_g^\top\mathbf{X}_g$. These terms only capture the second-order structure for their corresponding dataset. Consequently, $\mathbf{H}_\text{weighted}$ does not encode interactions between the two datasets.

We therefore propose a strategy which also incorporates such cross-dataset interactions. Concretely, we construct a mixed activation matrix $\mathbf{Z}$ that combines the activations from each of the calibration datasets. This is used to form the loss $\mathcal{L}_\text{mix}$:
\begin{equation*}
\mathbf{Z} = \sqrt{\alpha}\mathbf{X}_d + \sqrt{\beta}\mathbf{X}_g,\quad \mathcal{L}_\text{mix} = \mathcal{L}(\widehat{\mathbf{W}}, \mathbf{Z})
\end{equation*}
Expanding the Hessian for $\mathcal{L}_\text{mix}$ illustrates the additional cross term that is introduced:
\begin{align*}
\mathbf{H}_\text{mix} &= \mathbf{H}_\text{weighted} + 2\sqrt{\alpha\beta}(\mathbf{X}_d^\top\mathbf{X}_g + \mathbf{X}_g^\top\mathbf{X}_d)
\end{align*}
The contribution from this term can be interpreted as how strongly the domain-specific and general-purpose activations align.

Finally, we note a connection to the training-time data augmentation concept of \emph{mixup} \citep{zhang-etal-2018-mixup}, which forms linear combinations of pairs of examples and their labels. Mixup has previously been applied directly to activations \citep{verma-etal-2019-manifold} and Transformer LMs \citep{sun-etal-2020-mixup}. While our approach also forms linear combinations of activations, we use unlabeled examples solely to approximate the Hessian for compression.

\subsection{Implementation}

In practice, we compute the Hessian using the form $\mathbf{H}_\text{mix}=2\mathbf{Z}^\top\mathbf{Z}$. When the total number of calibration examples is held constant, this can reduce the cost of calibration, since each update to the Hessian incorporates two examples simultaneously. We analyze the resulting efficiency gains in \S\ref{sec:compression-efficiency}.

To empirically validate our approach, we integrate \ours{} with SparseGPT \citep{frantar-alistarh-2023-sparsegpt}, a popular Hessian-based pruning algorithm. However, we emphasize that \ours{} is not tied to any specific compression algorithm (see Figure~\ref{fig:quantization-results}, Appendix~\ref{app:quantization} for results with GPTQ-M).

\section{Experimental Setup}

\subsection{Compression Methods}
\label{sec:compression-methods}

Pruning methods can be formulated as a function that computes a saliency score $\mathbf{S}_{ij}$ for each weight $\mathbf{W}_{ij}$ in a given layer. They optionally use the layer input activations $\mathbf{X}$, derived from calibration data. We adopt the following methods as baselines.

\paragraph{Magnitude \normalfont{\citep{janowsky-1989-pruning, han-etal-2015-learning}.}}

Based on the assumption that removing the smallest weights will have the least effect, magnitude pruning simply uses the weight magnitude for saliency:
\begin{equation*}
\mathbf{S}_{ij} = |\mathbf{W}_{ij}|
\end{equation*}

\paragraph{SparseGPT \normalfont{\citep{frantar-alistarh-2023-sparsegpt}.}}

Building upon the OBS procedure, SparseGPT offers an efficient iterative approximation. The saliency metric is computed as follows, where $\lambda$ is a dampening factor to enable inversion of the Hessian:
\begin{equation*}
\mathbf{S}_{ij} = \left[ |\mathbf{W}|^2/\text{diag} \left((\mathbf{X}^\top\mathbf{X} + \lambda \mathbf{I})^{-1} \right) \right]_{ij}
\end{equation*}

\paragraph{Wanda \normalfont{\citep{sun-etal-2024-simple}.}}

Improving upon the computational efficiency of SparseGPT, the Wanda pruning metric approximates the diagonal of the inverse Hessian via the $\ell_2$ norm of the activations:
\begin{equation*}
\mathbf{S}_{ij} = |\mathbf{W}_{ij}| \cdot \|\mathbf{X}\|_2
\end{equation*}

\paragraph{D-Pruner \normalfont{\citep{zhang-etal-2024-pruning}.}} D-Pruner is a domain-specific pruning method. The first step of D-Pruner is to compute the general importance $\mathbf{G}_{ij}$ using a general dataset $\mathcal{D}_{g}$, similar to SparseGPT. Second, it uses a composite loss function $\mathcal{L}_\text{DP}$ to identify weights that are important for both general and domain-specific knowledge. This consists of the cross-entropy loss $\mathcal{L}_\text{CE}$ with a regularization term controlled by hyperparameter $\lambda_g$, where $\mathbf{W}'$ is the updated weight matrix:
\begin{equation*}
\mathcal{L}_\text{DP} \approx \mathcal{L}_\text{CE} + \lambda_g \sum_{i,j} \mathbf{G}_{ij} (\mathbf{W}'_{ij}-\mathbf{W}_{ij})^2
\end{equation*}
The gradients are computed using a full-parameter fine-tuning pipeline \citep{lv-etal-2024-full}. Given a domain-specific dataset $\mathcal{D}_d$, the saliency is:
\begin{equation*}
\mathbf{S}_{ij} \approx \left| \frac{\partial \mathcal{L}_\text{DP}(\mathcal{D}_d)}{\partial\mathbf{W}_{ij}}\mathbf{W}_{ij} + \frac{1}{2} \left[ \frac{\partial \mathcal{L}_\text{DP}(\mathcal{D}_d)}{\partial\mathbf{W}_{ij}}\mathbf{W}_{ij} \right]^2 \right|
\end{equation*}

\begin{algorithm}[t]
\small
\caption{\ours{} (simplified).}
\begin{algorithmic}[1]
\Require Domain dataset $\mathcal{D}_d$, generic dataset $\mathcal{D}_g$, weight $\alpha \in [0, 1]$, embedding matrix $\mathbf{U}$, and weight matrices $\mathcal{W}$.
\For{$i \gets 1$ to $|\mathcal{D}_d|$} \Comment{Since $|\mathcal{D}_d| = |\mathcal{D}_g|$.}
    \State $(\mathbf{X}_{d,i},\,\mathbf{X}_{g,i}) \gets (\mathcal{D}_{d,i} \cdot \mathbf{U},\, \mathcal{D}_{g,i} \cdot \mathbf{U})$
\EndFor

\For{$\mathbf{W} \in \mathcal{W}$}
    \State $\mathbf{H} \gets \mathbf{0}$

    \For{$i \gets 1$ to $|\mathcal{D}_d|$}
        \State $\mathbf{Z}_i \gets \sqrt{\alpha}\mathbf{X}_{d,i} + \sqrt{1 - \alpha}\mathbf{X}_{g,i}$
        \State $\mathbf{H} \gets \frac{i-1}{i}\mathbf{H} + \frac{2}{i}(\mathbf{Z}_i^\top\mathbf{Z}_i)$
    \EndFor
    \State $\mathbf{W} \gets \text{Compress}(\mathbf{W},\, \mathbf{H})$ \Comment{E.g. via SparseGPT.}
    \For{$i \gets 1$ to $|\mathcal{D}_d|$}
        \State $(\mathbf{X}_{d,i},\, \mathbf{X}_{g,i}) \gets (\mathbf{X}_{d,i} \mathbf{W},\, \mathbf{X}_{g,i} \mathbf{W})$
    \EndFor
\EndFor
\end{algorithmic}
\label{alg:ours}
\end{algorithm}

\begin{figure*}
\centering
\includegraphics{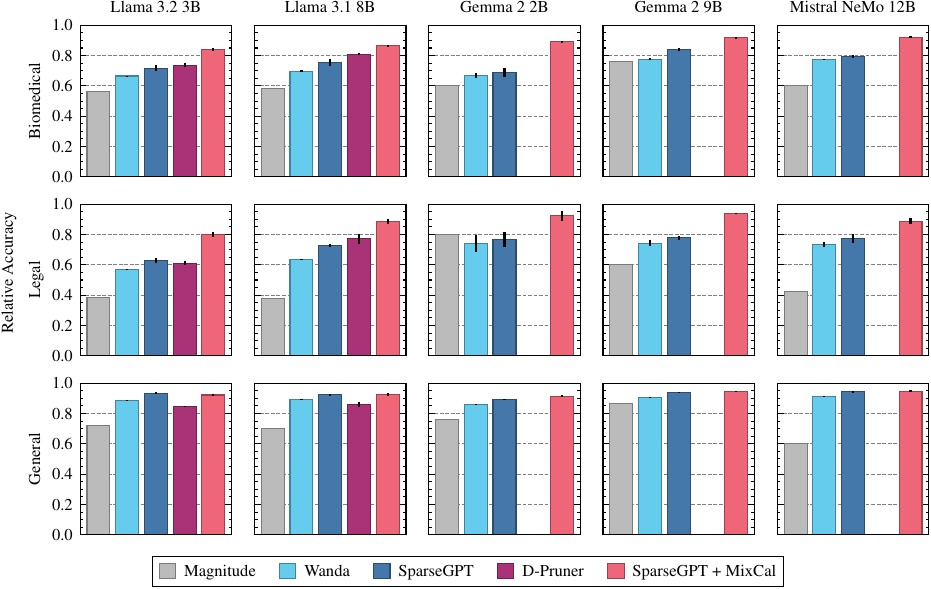}
\caption{The average benchmark accuracy when pruning to 50\% sparsity, relative to the original model.}
\label{fig:pruning-results}
\end{figure*}

\paragraph{GPTQ-M \normalfont{\citep{frantar-etal-2025-marlin}.}}

GPTQ adopts a Hessian-based weight sensitivity metric for quantization \citep{frantar-etal-2023-optq}. We select this method to enable a fair comparison with SparseGPT, which uses GPTQ for joint sparsification and quantization. We include improvements to the original algorithm suggested by \citet{frantar-etal-2025-marlin}. Specifically, this identifies optimal group-wise clipping thresholds, similar to AWQ \citep{lin-etal-2024-awq}. For clarity, we refer to this improved method as GPTQ-M.

\paragraph{Compression configurations.}

Guided by prior work \citep{sun-etal-2024-simple, frantar-etal-2025-marlin}, we focus our experiments on the following settings: 
\begin{itemize}[left=0pt]
    \item \textbf{50\% (unstructured) sparsity}. First, we experiment with individually pruning half of all layer weights, offering the highest possible granularity.
    \item \textbf{2:4 (semi-structured) sparsity}. We then examine pruning at the granularity of two weights in every group of four, enabling enhanced GPU inference performance \citep{mishra-etal-2021-accelerating}. 
    \item \textbf{4-bit quantization with 2:4 sparsity}. Finally, we combine 2:4 sparsity with 4-bit quantization of the remaining weights, enabling up to 5.3$\times$ GPU inference speedups \citep{frantar-etal-2025-marlin}.
\end{itemize}

\subsection{Domains and Tasks}
\label{sec:domains-and-tasks}

To assess the efficacy of our approach on downstream tasks, we experiment with two of the most widely explored domains in NLP, the \emph{biomedical} \citep{lee-etal-2019-biobert, gu-etal-2021-domain, luo-etal-2022-biogpt, singhal-etal-2023-large, singhal-etal-2025-toward} and \emph{legal} \citep{chalkidis-etal-2019-neural, zheng-etal-2021-when, henderson-etal-2022-pile, t-y-s-s-etal-2024-beyond, niklaus-etal-2024-multilegalpile} domains. See Appendix~\ref{app:datasets} for concrete task examples.

\paragraph{Biomedical.}

We use the MultiMedQA benchmark \citep{singhal-etal-2023-large}, specifically the PubMedQA \citep{jin-etal-2019-pubmedqa}, MedQA \citep{jin-etal-2021-disease}, MedMCQA \citep{pal-etal-2022-medmcqa} tasks, and relevant subsets from MMLU \citep{hendrycks-etal-2021-measuring} (anatomy, clinical knowledge, college medicine, medical genetics, professional medicine, college biology). To assess language modeling performance, we use the BioLaySumm PLOS dataset \citep{goldsack-etal-2022-making, goldsack-etal-2023-biolaysumm} comprising biomedical articles.

\paragraph{Legal.}

We follow \citet{colombo-etal-2024-saullm} in using Legal-MMLU, covering jurisprudence, professional law, and international law specialties \citep{hendrycks-etal-2021-measuring}. We also use the CaseHOLD \citep{zheng-etal-2021-when} and ECtHR (Task A) \citep{chalkidis-etal-2019-neural} datasets from the LexGLUE benchmark \citep{chalkidis-etal-2022-lexglue}, comprising US Supreme Court opinions and European Court of
Human Rights cases, respectively. To evaluate language modeling performance, we use the BillSum dataset \citep{kornilova-eidelman-2019-billsum} of US Congressional and California state bills.

\paragraph{General.}

To assess general performance, we use all commonsense reasoning tasks adopted by \citet{frantar-alistarh-2023-sparsegpt} and \citet{sun-etal-2024-simple}: ARC \citep{clark-etal-2018-think}, BoolQ \citep{clark-etal-2019-boolq}, HellaSwag \citep{zellers-etal-2019-hellaswag}, LAMBADA \citep{paperno-etal-2016-lambada}, OpenBookQA \citep{banerjee-etal-2019-careful}, PIQA \citep{bisk-etal-2020-piqa}, RTE \citep{dagan-etal-2006-pascal}, StoryCloze \citep{mostafazadeh-etal-2016-corpus}, and WinoGrande \citep{sakaguchi-etal-2021-winogrande}. We use WikiText-2 \citep{merity-etal-2017-pointer} to assess language modeling.

\subsection{Calibration Data}

\paragraph{Data sources.}

To create our general-purpose calibration sets, we follow \citet{dettmers-etal-2024-spqr} in using RedPajama \citep{weber-etal-2024-redpajama}, an open reproduction of the LLaMA training data. For the domain-specific calibration sets, we use MultiMedQA (biomedical) and LexGLUE (legal). In all cases, we sample data from the training splits only.

\paragraph{Data quantity.}

For a fair comparison between compression methods, we use 1024 calibration examples. As D-Pruner consists of two distinct stages to identify general and domain-specific weight importance, we allow 1024 examples from each dataset to better match the original work \citep{zhang-etal-2024-pruning}. For our own method, we simply select half of the examples (i.e. 512) from each dataset.

\paragraph{Sampling.}

We randomly sample segments of 2048 tokens following \citet{frantar-etal-2023-optq}, avoiding any selection bias. In the case of the domain-specific datasets, which may contain shorter examples, we follow \citet{touvron-etal-2023-llama2} in concatenating examples for a consistent length. We repeat the sampling process to create five distinct calibration sets, used to assess the variance in performance.

\subsection{Models}
\label{sec:models}

We experiment with popular open-weights LMs, covering different model families and sizes: (1) \textbf{Llama 3.2 3B} and \textbf{3.1 8B} \citep{grattafiori-etal-2024-llama3}, (2) \textbf{Gemma 2 2B} and \textbf{9B} \citep{riviere-etal-2024-gemma2}, and (3) \textbf{Mistral NeMo 12B} (2407) \citep{jiang-etal-2024-mistral}.

\section{Results \& Discussion}

\subsection{Pruning}
\label{sec:results-pruning}

Figure~\ref{fig:pruning-results} presents the benchmark accuracy when pruning to 50\% sparsity, relative to the original model.\footnote{The D-Pruner \citep{zhang-etal-2024-pruning} implementation only supports models with the Llama architecture. See \hyperref[sec:limitations]{Limitations}.} We report the mean value and standard deviation across five calibration sets. For brevity, we present the average performance across domain-specific models. We additionally present complete results across all models in Appendix~\ref{app:complete-results}.

\paragraph{A note on hyperparameters.}

To maximize the performance of the D-Pruner baseline, we perform an extensive hyperparameter search across $\lambda_g \in \{0.1, 0.01, 0.001\}$ and group size $\in \{\text{None}, 128\}$ for each model and domain. We then present results for only the best performing combinations. We present complete results across all hyperparameters in Appendix~\ref{app:complete-results}. 
In contrast, we do not optimize the hyperparameter for our approach ($\alpha$) and simply use 0.5 across all models and domains. We ablate the impact of this hyperparameter in Appendix~\ref{app:hyperparameters}.

\paragraph{\ours{} benefits domain performance.} 

We observe that across both biomedical and legal domains, our approach consistently outperforms all other compression methods. For example, we observe that \ours{} achieves an average relative accuracy of 88.6\% on the legal benchmark for Llama 3.1 8B. In comparison, SparseGPT and D-Pruner see 72.9\% and 77.2\%, respectively. For the biomedical benchmark, a similar trend can be observed. \ours{} achieves 86.5\%, while SparseGPT and D-Pruner reach 75.3\% and 81.0\%, respectively. This indicates that \ours{} can be effective at identifying weights influential to in-domain performance.

\begin{figure}[t]
\centering
\includegraphics{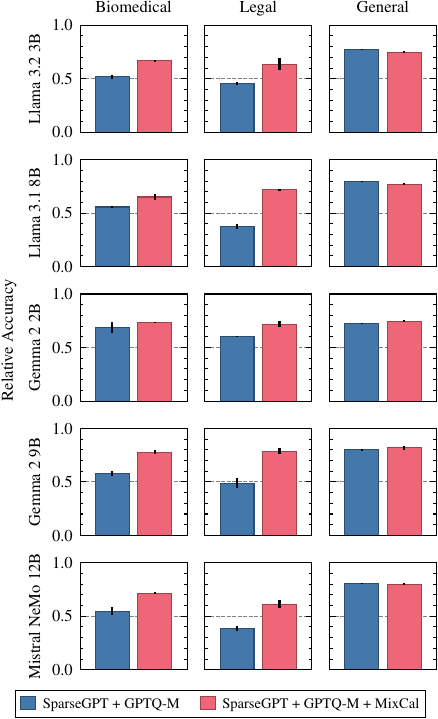}
\caption{Average accuracy when applying 4-bit quantization and 2:4 sparsity, relative to the original model.}
\label{fig:joint-results}
\end{figure}

\paragraph{General performance is comparable in all cases.}

In addition to substantial improvements on domain-specific benchmarks, general performance remains similar to the general-purpose pruning baselines. With Llama 3.1 8B, \ours{} achieves 92.6\% relative accuracy for general tasks on average. In comparison, SparseGPT achieves a slightly lower value of 92.2\%, while D-Pruner reaches only 85.7\%. 
Intriguingly, we note that \ours{} achieves higher performance than even the general-purpose methods for most model families. 
For example, \ours{} sees an average of 94.5\% with Gemma 2 9B, while SparseGPT reaches 93.7\%. This suggests that \ours{} can effectively establish domain-specific features without sacrificing general performance.

\paragraph{Performance gains are generally model-agnostic.}

Finally, we observe that the performance benefits of \ours{} are similar irrespective of the model size and family. For example, we consider two similarly sized models, Llama 3.1 8B and Gemma 2 9B. In the biomedical benchmark, we observe an 11.2 and 7.5 point increase in relative accuracy over SparseGPT, respectively. For the legal benchmark, we see a 15.7 and 16.1 point increase. This suggests that the performance gains from \ours{} are independent of the model family.

\paragraph{Language modeling follows a similar trend.}

Table~\ref{tab:complete-results-pruning} (Appendix~\ref{app:complete-results}) presents perplexity results. Similar to the downstream task experiments, we note that \ours{} achieves the best performance on in-domain language modeling. Considering Llama 3.1 8B, \ours{} achieves a perplexity of 6.0 compared to 6.7 with D-Pruner for the legal domain. For the biomedical domain, \ours{} has a perplexity of 10.7, compared to 14.7 from D-Pruner. The datasets used to evaluate perplexity are not used for calibration, suggesting that \ours{} assists with identifying \emph{domain}-specific features.

\paragraph{\ours{} appears to generalize beyond specific tasks.}

Across all models and domains, we observe that the performance benefits of \ours{} continue to tasks not included in the calibration data. In Table~\ref{tab:complete-results-analytical} (Appendix~\ref{app:complete-results}), we present complete per-task results. We consider the MMLU tasks, which do not have training data, and are therefore not represented in the calibration data. For Llama 3.1 8B, \ours{} achieves an absolute increase in accuracy of 7.7 points over SparseGPT in the biomedical domain. In the legal domain, \ours{} achieves an increase of 2.7 points in accuracy. This suggests that \ours{} can identify features that are relevant to the domain, rather than only a specific task.

\subsection{Joint Pruning and Quantization}
\label{sec:results_joint_compression}

We further examine the performance of our approach when jointly applying pruning and quantization by reusing the same inverse Hessian \citep{frantar-alistarh-2023-sparsegpt}. This has the advantage of allowing pruning and quantization decisions to influence each other, and enables quantization at almost no extra cost. Figure~\ref{fig:joint-results} presents benchmark accuracy when jointly applying 2:4 sparsity with 4-bit quantization, relative to the original model.

\paragraph{\ours{} improves in-domain performance while sustaining general performance.}

\ours{} achieves substantially greater domain performance than SparseGPT for 2:4 sparsity. For example, it sees a relative accuracy of 77.8\% on the biomedical benchmark for Gemma 2 9B, versus 58.2\% with SparseGPT. For general performance, we observe a similar trend to the pruning results, with \ours{} also performing comparably. This illustrates that \ours{} can be reliably used with quantization.

\subsection{Compression Efficiency}
\label{sec:compression-efficiency}

\begin{figure}[t]
\centering
\includegraphics{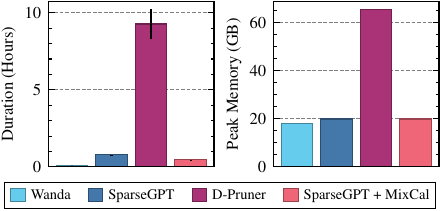}
\caption{The average duration and peak memory allocated when pruning Llama 3.1 8B with each method, as measured using an NVIDIA A100 80GB GPU.
}
\label{fig:runtime-statistics}
\end{figure}

\paragraph{\ours{} does not sacrifice efficiency.}

Figure~\ref{fig:runtime-statistics} presents the time and memory requirements of each method, which are often limiting factors in practice. 
First, we observe that \ours{} (0.5 hours) reduces the duration of compression compared to SparseGPT (0.8 hours). 
This is considerably faster than D-Pruner, which on average takes 9.3 hours, 18 times longer than our approach. We also observe that \ours{} does not increase the memory required for compression over SparseGPT, with both using up to 20 GB of memory. In contrast, D-Pruner uses up to 65.6 GB of memory, over three times more than our approach. This suggests that \ours{} is more practical than D-Pruner, offering lower computational resource requirements.

\section{Analysis}

\subsection{Ablating \ours{}}
\label{sec:ablation}

To better understand the role that in-domain data plays in pruning, we conduct an ablation study of \ours{}. Table~\ref{tab:results-ablation} presents the in-domain and general language modeling performance (perplexity) versus a SparseGPT baseline with generic calibration data. We specifically show the effect of (1) adding in-domain data to the calibration data mixture, and (2) additionally using \ours{}. In both cases, we use the same equal mixture of in-domain and generic data.
Results are averaged across models, with granular results in Table~\ref{tab:complete-results-ablation}, Appendix~\ref{app:complete-results}.

\begin{table}[t]
\scriptsize
\centering
\begin{tabular}{lccc}
\toprule
Method & In-domain & General & Average \\

\midrule
\multicolumn{4}{c}{Biomedical} \\ 
\midrule

SparseGPT & 16.4 & 14.0 & 15.2 \\
+ In-domain data & 15.1 & 13.8 & 14.4\\
+ \ours{} (Ours) & \textbf{14.4} & \textbf{13.4} & \textbf{13.9}\\
\midrule
\multicolumn{4}{c}{Legal}  \\
\midrule

SparseGPT & 7.2 & 14.0 & 10.6\\
+ In-domain data & \textbf{6.7} & 13.7 & 10.2\\
+ \ours{} (Ours) & \textbf{6.7} & \textbf{13.5} & \textbf{10.1}\\

\bottomrule
\end{tabular}
\caption{Perplexity when pruning to 50\% unstructured sparsity with (1) in-domain data, and (2) \ours{}, averaged across all models.}
\label{tab:results-ablation}
\end{table}

\paragraph{Domain-specific data benefits language modeling performance.}

We first observe that using a mix of domain-specific and generic calibration data can substantially improve in-domain performance. 
For example, when using a mix of in-domain data, we find lower perplexity scores for both domains. Specifically, we observe an average perplexity of 15.1 versus 16.4 for the biomedical domain, and 6.7 versus 7.2 for the legal domain. This highlights the influential role played by the calibration data, corroborating \citet{williams-aletras-2024-impact}.

\paragraph{\ours{} maximizes overall performance.}

Finally, we observe that by applying \ours{}, language modeling performance is maximized compared to introducing domain-specific calibration data alone. For example, \ours{} achieves a reduction of 0.7 (average) perplexity for the biomedical domain (14.4 versus 15.1). For the legal domain, we note that the average perplexity is equivalent. Considering overall performance, i.e. the average of domain-specific and general performance, \ours{} yields the best results. This suggests that the addition of \ours{} can balance domain-specific and general performance effectively.

\subsection{Performance in Other Languages}

To explore whether \ours{} generalizes beyond English, we further experiment with a Chinese-language model and benchmarks. We select the Chinese language as it is morphologically distinct from English and well-resourced in terms of models and domain-specific evaluation tasks.

\paragraph{Experimental setup.}

We select the Yi 1.5 6B model \citep{young-etal-2024-open} as it (1) uses the Llama architecture, enabling experiments with D-Pruner, and (2) achieves strong performance on standard benchmarks. To assess biomedical performance, we use the Comprehensive Medical Benchmark (CMB; \citealp{wang-etal-2024-cmb}). For legal performance, we use the Chinese AI
and Law (CAIL2018) challenge dataset \citep{xiao-etal-2018-cail2018}. Similar to our experimental setup in English, we sample in-domain calibration data from the training split of each benchmark. For general-purpose data, we follow \citet{kurz-etal-2026-limitations} in using mC4 \citep{xue-etal-2021-mt5}. We use a mixture of Chinese and English to reflect the model pre-training data \citep{young-etal-2024-open}.

\begin{figure}[t]
\centering
\includegraphics{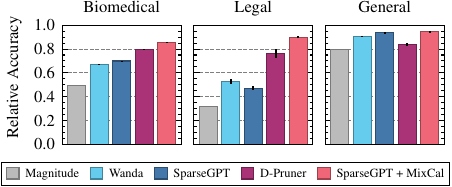}
\caption{The mean accuracy for Yi 1.5 6B on Chinese-language benchmarks, relative to the original model.}
\label{fig:results-pruning-zh}
\end{figure}

\paragraph{\ours{} appears language-agnostic.}

Figure~\ref{fig:results-pruning-zh} presents the relative benchmark accuracy when pruning to 50\% sparsity, similar to earlier experiments (\S \ref{sec:results-pruning}).
The results indicate that \ours{} outperforms other pruning approaches for domain performance, while maintaining comparable general performance to SparseGPT. 
For example, \ours{} achieves a relative accuracy of 85.6\% in the biomedical domain, compared to 70.0\% and 79.6\% from SparseGPT and D-Pruner, respectively.
In the legal domain, the performance retention from \ours{} (90.1\%) is substantially greater than both SparseGPT (47.1\%) and D-Pruner (76.2\%).
These findings are in line with the English language tasks, suggesting that \ours{} is language-agnostic.

\section{Conclusion}

In this paper, we proposed \ours{} as a solution for creating compressed LMs for specialized domains. We empirically validated \ours{} using a plethora of pre-trained models and evaluation tasks. Our approach represents a substantial advancement over earlier work such as D-Pruner, offering consistent performance improvements with a smaller computational footprint. We hope that our study will inspire further work towards the efficient deployment of LMs in specialized domains. As future work, we are interested in exploring how synthetic calibration data could be incorporated to further enhance the performance of domain-specific LM compression \citep{williams-etal-2025-self}.

\section*{Limitations}
\label{sec:limitations}

\paragraph{Model selection in the D-Pruner experiments.}

The D-Pruner \citep{zhang-etal-2024-pruning} implementation supports only the Llama model architecture.\footnote{\url{https://github.com/psunlpgroup/D-Pruner}} Consequently, we are limited to offering comparisons for only the models using the Llama architecture (i.e. Llama 3.2 3B, Yi 1.5 6B, and Llama 3.1 8B). We emphasize that our approach substantially outperforms D-Pruner across all tested models and domains. Therefore, we expect that this trend would continue for the models not supported by D-Pruner.

\section*{Ethical Considerations}

Our work enables the efficient and effective compression of LMs for specialized domains. We note that this poses dual-use concerns, as it may enable misuse at a lower cost \citep{weidinger-etal-2022-taxonomy}. However, we emphasize that our approach is unlikely to enhance or introduce new harmful abilities, as the performance of compressed models is constrained by the capabilities of the original.

\section*{Acknowledgments}

We would like to thank Huiyin Xue and the anonymous reviewers for their invaluable feedback. Additionally, we are sincerely grateful to Huiyin Xue for assistance with the Chinese-language tasks. MW is supported by the Centre for Doctoral Training in Speech and Language Technologies (SLT) and their Applications funded by UK Research and Innovation grant EP/S023062/1. NA is supported by EPSRC grant EP/Y009800/1, part of the RAI UK Keystone projects.

\bibliography{anthology-1,custom}

\appendix

\section{Additional Ablation Studies}

\subsection{4-bit Quantization}
\label{app:quantization}

\paragraph{\ours{} generalizes to quantization.}

\begin{figure}[t]
\centering
\includegraphics{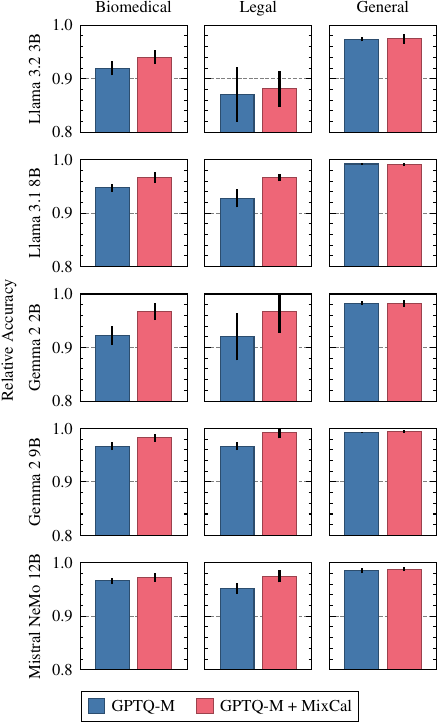}
\caption{Average accuracy when applying 4-bit quantization, relative to the original model.}
\label{fig:quantization-results}
\end{figure}

We further examine the performance and transferability of our approach under the setting of quantization. Figure~\ref{fig:quantization-results} presents the benchmark accuracy when applying 4-bit quantization with GPTQ-M and \ours{}, relative to the original model. This uncovers the extent to which \ours{} benefits each method independently in the joint sparsification and quantization experiments (\S\ref{sec:results_joint_compression}).
We observe that general performance is consistently preserved, with GPTQ-M + \ours{} performing comparably to GPTQ-M across all models. This suggests that \ours{} is  transferable across settings in Hessian-based compression.

\paragraph{\ours{} improves in-domain performance.}

We observe that \ours{} can achieve greater domain performance than GPTQ-M alone. For example, \ours{} sees a relative accuracy of 96.6\% on the biomedical benchmark for Llama 3.1 8B, versus 94.7\% with GPTQ-M (Figure~\ref{fig:quantization-results}). This indicates that \ours{} can be reliably used for LM quantization. We note that the performance gains for quantization are smaller than in the pruning experiments. This is expected, as quantization is less sensitive to calibration data at the tested sparsity level and bit width \citep{williams-aletras-2024-impact}.

\subsection{Performance Across Sparsity Levels}

\begin{figure*}[p]
\centering
\includegraphics{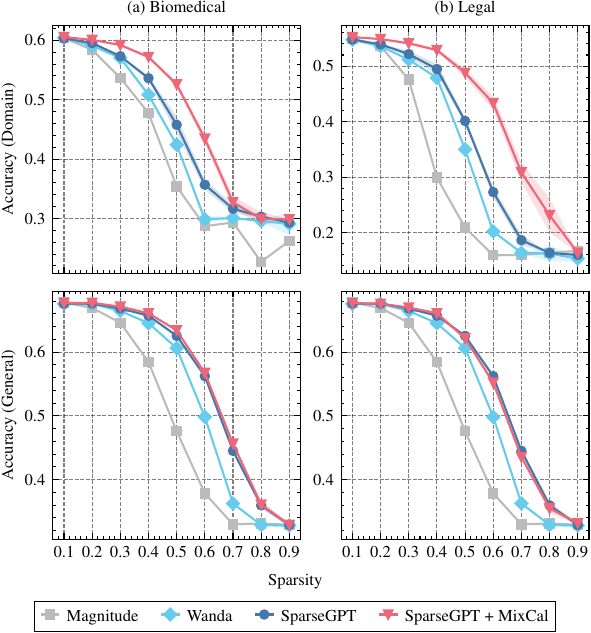}
\caption{Benchmark accuracy when pruning Llama 3.1 8B to different sparsity levels with each method. Standard deviation is denoted by the shaded regions.}\label{fig:ablation-sparsity}
\end{figure*}

In Figure~\ref{fig:ablation-sparsity}, we examine how \ours{} performs compared to the other pruning methods across different sparsity levels. We report the average in-domain and general benchmark performance, with standard deviation denoted by the shaded regions. We first observe that in-domain performance with \ours{} consistently remains higher than other approaches, even beyond 50\% sparsity. For example, accuracy on the legal benchmark at 60\% sparsity is 43.3\% with \ours{}, compared to 27.3\% with SparseGPT. Similarly, general benchmark performance remains comparable to SparseGPT across all sparsity levels. This suggests that \ours{} can be used to effectively isolate domain-specific features without sacrificing generic performance.

\subsection{\ours{} Hyperparameter Analysis}
\label{app:hyperparameters}

\begin{table}[t]
\scriptsize
\centering
\begin{tabular}{llc}
\toprule
Method & Hyperparameter & Value \\
\midrule
\multirow{3}{*}{D-Pruner} & Loss Regularization & \{0.001, 0.01, 0.1\} \\ 
& Group Size & \{None, 128\} \\
& Learning Rate & 0.03 \\
\midrule
\multirow{4}{*}{GPTQ-M}  & Bits per Weight & 4 \\
& Dampening & 0.01 \\
& Group Size & 128 \\
& Symmetric Quantization & Yes \\
\midrule
\multirow{3}{*}{SparseGPT} & Dampening & 0.01 \\
& Group Size & 128 \\
& Sparsity & \{0.5, 2:4\} \\
\midrule
\multirow{2}{*}{Wanda} & Group Size & 1 \\
& Sparsity & \{0.5, 2:4\} \\
\bottomrule
\end{tabular}
\caption{Hyperparameters used for all compression methods evaluated in our experiments.}
\label{tab:hyperparameters}
\end{table}

\begin{figure*}[p]
\centering
\includegraphics{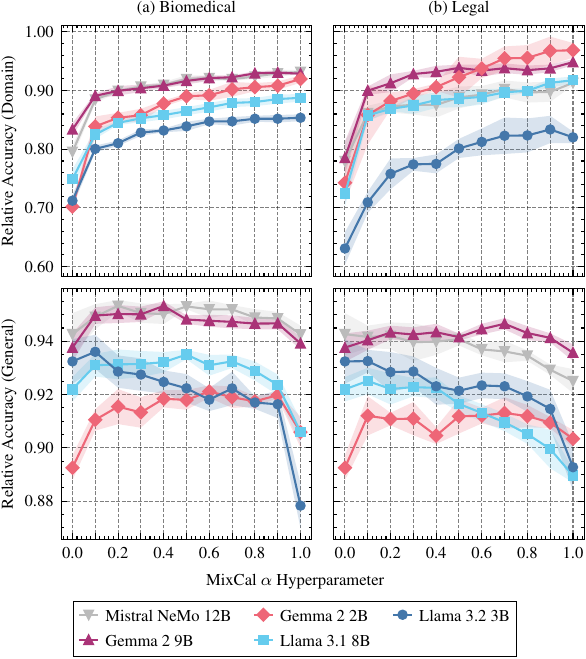}
\caption{Benchmark accuracy relative to the original model when varying the \ours{} $\alpha$ hyperparameter for pruning. Standard deviation is denoted by the shaded regions.}\label{fig:ablation-alpha}
\end{figure*}

In Figure~\ref{fig:ablation-alpha}, we examine how different contributions from domain-specific and general-purpose data impact benchmark performance, i.e. by varying $\alpha$. Following the experiments in \S\ref{sec:results-pruning}, we examine this at 50\% unstructured sparsity.

\paragraph{Increasing $\boldsymbol{\alpha}$ generally increases in-domain performance.}

First, we observe that in-domain performance (i.e. biomedical or legal) increases following a larger contribution from the domain-specific Hessian. Considering Gemma 2 9B for the biomedical domain, we observe that $\alpha=0.1$ achieves an average accuracy of 89.1\%, whilst $\alpha=0.9$ sees 93.1\%. Notably, we observe that even a small contribution from the in-domain Hessian ($\alpha=0.1$) is enough to offer improvements for in-domain performance. For example, in the legal domain Llama 3.2 3B achieves an average accuracy of 63.1\% with $\alpha=0$, yet 71.0\% with $\alpha=0.1$.

\paragraph{Maximizing $\alpha$ can harm reasoning performance.}

We observe that using $\alpha = 1$ can lead to a drop in reasoning performance. For example, when targeting the legal domain, Llama 3.2 3B achieves an average accuracy of 91.5\% with $\alpha = 0.9$, yet sees 89.3\% when $\alpha = 1$. This suggests that the use of generic data can benefit reasoning performance.

\section{Hyperparameters}

Table~\ref{tab:hyperparameters} presents the hyperparameters of all the compression methods in our experiments. In general, we adopt the optimal hyperparameters used for each method in the original work. For completeness, we also present results for every tested D-Pruner hyperparameter combination in Table~\ref{tab:complete-results-dpruner}.

\section{Data \& Processing}
\label{app:datasets}

Table~\ref{tab:task-statistics} shows the splits for the datasets used in our experiments, split by category. We also show examples of how the domain-specific tasks are formatted for zero-shot LM evaluation. These are presented in Tables \ref{tab:task-examples-legal} and \ref{tab:task-examples-biomedical} for the legal and biomedical domains, respectively. 

In the majority of cases, we use the task datasets exactly as implemented by the EleutherAI LM Evaluation Harness \citep{gao-etal-2024-framework}. We highlight the exceptions where additional preprocessing was required, below:

\begin{itemize}[left=0pt]
\item \textbf{CaseHOLD} and \textbf{ECtHR (Task A)}. We adopt the versions of these datasets as provided by the LexGLUE benchmark \citep{chalkidis-etal-2022-lexglue}. To enable evaluation in a zero-shot setting, we adapt the prompts from \citet{chalkidis-etal-2023-chatgpt}.
\item \textbf{ECtHR (Task A)}, \textbf{CAIL2018}, and \textbf{CMB}. We additionally filter examples with multiple labels from these datasets, following prior work towards adapting existing tasks for few-shot LM evaluation \citep{guha-etal-2023-legalbench}.
\item \textbf{CAIL2018} and \textbf{CMB}. For Chinese-language evaluation tasks, we use the prompts presented in Table~\ref{tab:task-examples-zh}, Appendix~\ref{app:datasets}.
\item \textbf{BillSum} and \textbf{BioLaySumm PLOS}. Due to the extensive size of the test split in these datasets, we follow \citet{frantar-etal-2023-optq} in using the first 256 examples to assess perplexity. We highlight that perplexity is a stable metric which can be assessed using only a small number of examples \citep{dettmers-etal-2023-case}.
\end{itemize}

\begin{table}[t]
\scriptsize
\centering
\begin{tabular}{lrrr}
\toprule
Name                         & Train  & Val.   & Test \\ \midrule
\multicolumn{4}{c}{Biomedical} \\ \midrule
BioLaySumm PLOS             & 24,773  & 1,376  & 256  \\
CMB (Chinese)                        & 231,902 &  228 & 9,325  \\
PubMedQA                     & 450    & 50    & 500  \\
MedMCQA                      & 182,822 & 4,183  & 4,183 \\
MedQA                        & 10,178  & 1,272  & 1,273 \\
MMLU Anatomy                &        &       & 135  \\
MMLU Clinical Knowledge    &        &       & 265  \\
MMLU College Medicine      &        &       & 173  \\
MMLU Medical Genetics      &        &       & 100  \\
MMLU Professional Medicine &        &       & 272  \\
MMLU College Biology       &        &       & 144  \\ \midrule
\multicolumn{4}{c}{Legal} \\ \midrule
BillSum                      & 18,949  &       & 256  \\
CAIL2018 (Chinese)                   & 110,905 & 14,147 & 27,484 \\
MMLU International Law     &        &       & 121  \\
MMLU Jurisprudence          &        &       & 108  \\
MMLU Professional Law      &        &       & 1,534 \\
LexGLUE CaseHOLD            & 45,000  & 3,900  & 3,600 \\
LexGLUE ECtHR (Task A)      & 6,838   & 802  & 808 \\ \midrule
\multicolumn{4}{c}{General} \\ \midrule
ARC (Easy)                    & 2,251   & 570   & 2,376 \\
ARC (Challenge)               & 1,119   & 299   & 1,172 \\
BoolQ                        & 9,427   & 3,270  &      \\
HellaSwag                    & 39,905  & 10,042 &      \\
LAMBADA (Standard)            &        & 4,869  & 5,153 \\
OpenBookQA                   & 4,957   & 500   & 500  \\
PIQA                         & 16,113  & 1,838  &      \\
RTE                          & 2,490   & 277   &      \\
WinoGrande                   & 40,398  & 1,267  &      \\
StoryCloze                 & 360    & 1,511  &      \\ \bottomrule
\end{tabular}
\caption{Number of examples in each evaluation task.}
\label{tab:task-statistics}
\end{table}

\section{Infrastructure}

We implement all experiments using PyTorch \citep{paszke-etal-2019-pytorch} with the model implementations from Hugging Face Transformers \citep{wolf-etal-2020-transformers}. We additionally use Hugging Face Datasets \citep{lhoest-etal-2021-datasets} for all dataset manipulation, including for tasks implemented via the EleutherAI LM Evaluation Harness \citep{gao-etal-2024-framework}. Finally, we conduct all experiments using a single NVIDIA A100 (SXM4 80GB) GPU.

\begin{table*}[t]
    \small
    \centering
    \begin{tabular}{p{2cm}|p{10cm}p{2cm}}
        \toprule
        Task & Input & Answer \\ \midrule
         
        \multirow{18}{*}{CaseHOLD} & \textit{Given the following excerpt from a United States court opinion:} & \multirow{18}{*}{Holding B} \\
         & & \\
          & \textbf{Citing Text:} & \\
          & ... Warner-Lambert Co., 427 Mass. at 49 (“ [Confidential and proprietary business information may be entitled to protection, even if such information cannot claim trade secret protection”); see, e.g., Augat, Inc., 409 Mass. at 173 (\underline{<HOLDING>}). “Matters of public knowledge or of general... \\
         & \textit{Given the following excerpt from a United States court opinion:} \\ 
         & \\ 
         & \textit{ Which one of the following options should replace the <HOLDING> placeholder?}\\
         & \\
         & \textbf{Holdings:} \\ 
         & A. Recognizing that even if a plaintiff claims certain information constitutes trade secrets its claim may not depend on that  determination. \\
          & B. Holding that included among trade secrets employee may not appropriate from employer is certain information such as lists of customers. \\ 
        & C. ... & \\ & \\ \midrule
         
         \multirow{15}{*}{ECtHR Task A} & \textit{Given the following facts from a European Court of Human Rights (ECtHR) case:} & \multirow{15}{*}{Choice K} \\
         & & \\
        & \textbf{Articles: } \\ 
        & 2. On 8 May 1996 the applicant was arrested in New York (USA) and placed in detention on the basis of a extradition request  from the authorities of the Netherlands Antilles where ... \\
        & 3. In a document dated 18 June 1996 bearing the applicant’s ... \\ & \textit{Which article of the European Convention on Human Rights (ECHR) has been violated?} \\ 
        & \textit{A. Article 2} \\ 
        & \textit{B. Article 3} \\ 
        & \textit{C. Article 5} \\
        & ... \\
        & \textit{K. None of the above} & \\ & \\ \midrule

        \multirow{10}{*}{Jurisprudence} & \textit{The following are multiple choice questions (with answers) about jurisprudence:} & \multirow{10}{*}{Choice B} \\
         & & \\
        & \textbf{Question: } \\ 
      &  Which statement best explains the purpose of Hart's distinction between 'being obliged' and 'having an obligation'? \\ 
       & \\
         & \textbf{Choices:} \\ 
        & A. It demonstrates the difference between the internal and the external aspect of a rule.\\ 
        & B. It refutes the natural lawyer' ... \\ 
        & C. ... & \\ & \\ \midrule

        \multirow{10}{*}{\makecell[l]{International / \\ Professional \\ Law}} & \textit{The following are multiple choice questions (with answers) about professional / international law:} & \multirow{10}{*}{Choice B} \\
         & & \\
        & \textbf{Statement: } \\ 
      &  One afternoon, a pilot was flying a small airplane when it suddenly ran out of gas. As he was coming in for an emergency landing ... The attorney's testimony is: \\ 
       & \\
         & \textbf{Choices:} \\ 
        & A. admissible, because the...\\ 
        & B. inadmissible because ... \\ 
        & C. ... & \\ & \\ \bottomrule
         
    \end{tabular}
    \caption{Representative task examples from the legal benchmark. \textit{Italicized} text denotes the prompt.}
    \label{tab:task-examples-legal}
\end{table*}

\begin{table*}[t]
    \small
    \centering
    \begin{tabular}{p{2cm}|p{10cm}p{2cm}}
        \toprule
        Task & Input & Answer \\ \midrule
         
        \multirow{10}{*}{\makecell[l]{\textbf{MMLU Tasks}: \\ Clinical \\ Knowledge; \\ \\ College \\ Medicine; \\ \\ College \\ Biology; \\ \\ Professional \\ Medicine; \\ \\ Anatomy;  \\ \\ Medical \\ Genetics. }} & \textit{The following are multiple choice questions (with answers) about <TASK>.} & \multirow{18}{*}{Choice A} \\
         & & \\
          & \textbf{Question:} & \\
          & What size of cannula would you use in a patient who needed a rapid blood transfusion (as of 2020 medical knowledge)? \\ 
         & \\ 
         & \textbf{Choices:} \\ 
         & A. 18 gauge \\
         & B. 20 gauge \\
         & C. ... \\
         & \\
         & \\
         & \\& \\ & \\ & \\ & \\
         & & \\ \midrule

         \multirow{14}{*}{\makecell[l]{PubMedQA}} & \textit{Abstract} & \multirow{14}{*}{Choice A - Yes} \\
          & To evaluate the degree to which histologic chorioamnionitis, a frequent finding in placentas submitted for histopathologic evaluation, correlates with clinical indicators of infection in the mother. A retrospective review was performed on 52 cases with a histologic diagnosis of acute chorioamnionitis from 2,051 ... \\ 
         & \\ 
         & \textit{Question} & \\
         & Does histologic chorioamnionitis correspond to clinical chorioamnionitis? \\
        & \\
         & \textit{Choices:} \\ 
         & A. Yes \\
         & B. No \\
         & C. Maybe \\  & \\ \midrule

         \multirow{9}{*}{\makecell[l]{MedMCQA}} & \textit{Question:}  & \multirow{9}{*}{Choice D} \\
          & All of the following are surgical options for morbid obesity except -: \\ 
         & \\ 
         & \textit{Choices:} \\ 
         & A. Adjustable gastric banding  \\
         & B. Biliopancreatic diversion \\
         & C. Duodenal Switch \\
         & D. ... \\ & \\ \midrule

         \multirow{10}{*}{\makecell[l]{MedQA}} & \textit{Question:} & \multirow{10}{*}{Choice A} \\
          & A 5-year-old girl is brought to the clinic by her mother for excessive hair growth. Her mother reports that for the past 2 months she has noticed ... studies demonstrates an elevated level of estrogen. What is the most likely diagnosis? \\ 
         & \\ 
         & \textit{Choices:} \\ 
         & A. Granulosa cell tumor  \\
         & B. Idiopathic precocious puberty \\
         & C. ... \\ & \\ \bottomrule

    \end{tabular}
    \caption{Representative task examples from the biomedical benchmark. \textit{Italicized} text denotes the prompt.}
    \label{tab:task-examples-biomedical}
\end{table*}

\begin{table*}[t]
    \small
    \centering
    \begin{tabular}{p{2cm}|p{10cm}p{2cm}}
        \toprule
        Task & Input & Answer \\ \midrule
         
        \multirow{12}{*}{CAIL2018} & \chinese{根据以下法律案件的事实:} & \multirow{12}{*}{Choice A} \\
         & & \\
          & <FACT> & \\
         & \\ 
         & \chinese{请问中华人民共和国刑法中的哪一条适用于本案?} \\ 
         & A. \chinese{第} <ARTICLE 1> \chinese{条} \\
         & B. \chinese{第} <ARTICLE 2> \chinese{条} \\
         & C. \chinese{第} <ARTICLE 3> \chinese{条} \\
         & D. \chinese{第} <ARTICLE 4> \chinese{条} \\
         & \\
         & \chinese{答案:} \\
         & & \\ \midrule

          \multirow{10}{*}{CMB} & \chinese{问题}: <QUESTION> & \multirow{10}{*}{Choice A} \\
         & & \\
        &  A. <OPTION 1> \\
        &  B. <OPTION 2> \\
        &  C. <OPTION 3> \\
        &  D. <OPTION 4> \\
        &  E. <OPTION 5> \\
         & \\
         & \chinese{答案:} \\
         & & \\ \bottomrule

    \end{tabular}
    \caption{Example format of tasks in Chinese-language evaluation (CAIL2018 for legal and CMB for biomedical).}
    \label{tab:task-examples-zh}
\end{table*}

\section{Complete Results}
\label{app:complete-results}

Table~\ref{tab:complete-results-pruning} presents the full benchmark results (accuracy and perplexity) corresponding to Figure~\ref{fig:pruning-results}. For D-Pruner, we present results using iterative blocking and a loss regularization of 0.001.  Finally, in Table~\ref{tab:complete-results-analytical} we further decompose benchmark results into their constituent tasks for completeness.

\begin{table*}[t]
    \centering
    \scriptsize
    \begin{tabular}{lllcc|cc|cc}
\toprule

\multirow{2}{*}{Model} & \multirow{2}{*}{Method} & Target & \multicolumn{2}{c|}{General} & \multicolumn{2}{c|}{Legal} & \multicolumn{2}{c}{Biomedical} \\
& & Domain & Accuracy & Perplexity & Accuracy & Perplexity & Accuracy & Perplexity \\
\midrule

\multirow{8}{*}{Llama 3.2 3B}& - & - & \referenceresult{61.6}{0.0} & \referenceresult{9.3}{0.0} & \referenceresult{45.5}{0.0} & \referenceresult{5.2}{0.0} & \referenceresult{53.3}{0.0} & \referenceresult{9.4}{0.0} \\

 & Magnitude & - & \result{44.2}{0.0} & \result{50.7}{0.0} & \result{17.4}{0.0} & \result{24.9}{0.0} & \result{30.1}{0.0} & \result{62.2}{0.0} \\
 & Wanda & - & \result{54.4}{0.2} & \result{15.4}{0.1} & \result{26.0}{0.2} & \result{8.9}{0.0} & \result{35.5}{0.1} & \result{15.5}{0.0} \\
 & SparseGPT & - & \bestresult{57.5}{0.4} & \result{13.8}{0.1} & \result{28.6}{0.7} & \result{8.2}{0.1} & \result{38.2}{1.1} & \result{14.7}{0.1} \\ \cmidrule(lr){2-9}
 
 &  \multirow{2}{*}{D-Pruner} & Legal & \result{52.2}{0.2} & \result{16.0}{0.1} & \result{27.8}{0.6} & \result{8.2}{0.0} & - & - \\
 & & Biomedical & \result{52.1}{0.1} & \result{19.0}{0.2} & - & - & \result{39.0}{0.6} & \result{16.3}{0.1} \\ \cmidrule(lr){2-9}
 
 & \multirow{2}{*}{SparseGPT + \ours{}} & Legal & \result{56.8}{0.3} & \result{13.6}{0.0} & \bestresult{36.5}{0.7} & \bestresult{7.7}{0.2} & - & - \\
 &  & Biomedical & \result{56.8}{0.3} & \bestresult{13.3}{0.0} & - & - & \bestresult{44.7}{0.4} & \bestresult{13.5}{0.1} \\ \midrule

\multirow{8}{*}{Llama 3.1 8B} & - & - & \referenceresult{67.8}{0.0} & \referenceresult{7.3}{0.0} & \referenceresult{55.1}{0.0} & \referenceresult{4.2}{0.0} & \referenceresult{60.8}{0.0} & \referenceresult{8.0}{0.0} \\

& Magnitude & - & \result{47.6}{0.0} & \result{57.7}{0.0} & \result{20.9}{0.0} & \result{65.7}{0.0} & \result{35.4}{0.0} & \result{55.7}{0.0} \\
 & Wanda & - & \result{60.6}{0.1} & \result{11.7}{0.0} & \result{35.0}{0.3} & \result{6.8}{0.0} & \result{42.4}{0.2} & \result{11.9}{0.0} \\
& SparseGPT & - & \result{62.6}{0.2} & \result{10.7}{0.1} & \result{40.1}{0.6} & \result{6.3}{0.1} & \result{45.8}{1.4} & \result{11.5}{0.1} \\ \cmidrule(lr){2-9}

 & \multirow{2}{*}{D-Pruner}  & Legal & \result{57.2}{0.3} & \result{12.6}{0.2} & \result{42.5}{1.8} & \result{6.7}{0.0} & - & - \\
&  & Biomedical & \result{59.1}{0.2} & \result{13.5}{0.1} & - & - & \result{49.2}{0.4} & \result{14.7}{0.3} \\ \cmidrule(lr){2-9}

 & \multirow{2}{*}{SparseGPT + \ours{}} & Legal & \result{62.2}{0.2} & \result{10.7}{0.0} & \bestresult{48.8}{0.8} & \bestresult{6.0}{0.1} & - & - \\
 &  & Biomedical & \bestresult{63.4}{0.1} & \bestresult{10.5}{0.0} & - & - & \bestresult{52.6}{0.4} & \bestresult{10.7}{0.0} \\ \midrule

\multirow{6}{*}{Gemma 2 2B} & - & - & \referenceresult{63.3}{0.0} & \referenceresult{13.1}{0.0} & \referenceresult{27.1}{0.0} & \referenceresult{6.2}{0.0} & \referenceresult{44.2}{0.0} & \referenceresult{15.0}{0.0} \\
 & Magnitude & - & \result{48.2}{0.0} & \result{172.5}{0.0} & \result{21.7}{0.0} & \result{34.7}{0.0} & \result{26.8}{0.0} & \result{825.8}{0.0} \\
 & Wanda & - & \result{54.2}{0.2} & \result{25.0}{0.2} & \result{20.1}{1.5} & \result{10.7}{0.0} & \result{29.6}{0.7} & \result{33.0}{0.5} \\
 & SparseGPT & - & \result{56.6}{0.2} & \result{20.9}{0.4} & \result{20.8}{1.4} & \result{9.2}{0.1} & \result{30.4}{1.3} & \result{26.6}{0.3} \\ \cmidrule(lr){2-9}

 & \multirow{2}{*}{SparseGPT + \ours{}} & Legal & \result{57.7}{0.4} & \bestresult{18.9}{0.1} & \bestresult{25.0}{0.9} & \bestresult{8.3}{0.0} & - & - \\
 &  & Biomedical & \bestresult{58.1}{0.2} & \result{19.1}{0.1} & - & - & \bestresult{39.4}{0.2} & \bestresult{21.5}{0.2} \\ \midrule

\multirow{6}{*}{Gemma 2 9B} & - & - & \referenceresult{70.2}{0.0} & \referenceresult{10.6}{0.0} & \referenceresult{56.2}{0.0} & \referenceresult{4.8}{0.0} & \referenceresult{62.8}{0.0} & \referenceresult{12.0}{0.0} \\
 & Magnitude & - & \result{60.6}{0.0} & \result{33.5}{0.0} & \result{33.7}{0.0} & \result{11.0}{0.0} & \result{47.8}{0.0} & \result{57.4}{0.0} \\
 & Wanda & - & \result{63.5}{0.2} & \result{16.6}{0.1} & \result{41.8}{1.2} & \result{6.7}{0.0} & \result{48.8}{0.4} & \result{20.8}{0.2} \\
 & SparseGPT & - & \result{65.8}{0.2} & \result{15.2}{0.2} & \result{43.7}{0.8} & \result{6.4}{0.0} & \result{52.8}{0.5} & \result{19.3}{0.2} \\ \cmidrule(lr){2-9}
 
 &\multirow{2}{*}{SparseGPT + \ours{}}& Legal & \result{66.1}{0.1} & \result{14.6}{0.0} & \bestresult{52.8}{0.3} & \bestresult{5.9}{0.0} & - & - \\
 &  & Biomedical & \bestresult{66.5}{0.1} & \bestresult{14.5}{0.1} & - & - & \bestresult{57.6}{0.4} & \bestresult{16.8}{0.1} \\ \midrule

\multirow{6}{*}{Mistral NeMo 12B} & - & - & \referenceresult{69.4}{0.0} & \referenceresult{7.1}{0.0} & \referenceresult{57.5}{0.0} & \referenceresult{4.3}{0.0} & \referenceresult{58.4}{0.0} & \referenceresult{7.6}{0.0} \\
 & Magnitude & - & \result{41.8}{0.0} & \result{465.5}{0.0} & \result{24.5}{0.0} & \result{3.6$\times10^3$}{0.0} & \result{35.0}{0.0} & \result{10.3$\times10^3$}{0.0} \\
 & Wanda & - & \result{63.2}{0.2} & \result{10.3}{0.0} & \result{42.3}{0.8} & \result{6.0}{0.0} & \result{45.3}{0.3} & \result{10.6}{0.0} \\
 & SparseGPT & - & \result{65.4}{0.4} & \bestresult{9.4}{0.0} & \result{44.5}{1.8} & \result{5.6}{0.0} & \result{46.4}{0.6} & \result{9.8}{0.0} \\ \cmidrule(lr){2-9}
 &\multirow{2}{*}{SparseGPT + \ours{}}& Legal & \result{65.3}{0.2} & \result{9.5}{0.0} & \bestresult{51.1}{1.1} & \bestresult{5.5}{0.0} & - & - \\
 &  & Biomedical & \bestresult{66.2}{0.1} & \bestresult{9.4}{0.0} & - & - & \bestresult{53.7}{0.4} & \bestresult{9.6}{0.0} \\ \bottomrule
\end{tabular}

    \caption{Average performance for the general and domain-specific benchmarks when pruning with 50\% sparsity. Standard deviations are shown as subscripts. \textbf{Bold} values denote the best performing method for each model. For reference, the top row of each model shows the original model performance (i.e. prior to compression).}
    \label{tab:complete-results-pruning}
\end{table*}

\begin{table*}[t]
    \centering
    \scriptsize
    \begin{tabular}{lllcc|cc|cc}
\toprule

\multirow{2}{*}{Model} & \multirow{2}{*}{Method} & Target & \multicolumn{2}{c|}{General} & \multicolumn{2}{c|}{Legal} & \multicolumn{2}{c}{Biomedical} \\
& & Domain & Accuracy & Perplexity & Accuracy & Perplexity & Accuracy & Perplexity \\
\midrule

\multirow{7}{*}{Llama 3.2 3B} & - & - & \referenceresult{61.6}{0.0} & \referenceresult{9.3}{0.0} & \referenceresult{45.5}{0.0} & \referenceresult{5.2}{0.0} & \referenceresult{53.3}{0.0} & \referenceresult{9.4}{0.0} \\

 & SparseGPT & - & \result{49.6}{0.2} & \bestresult{22.6}{0.2} & \result{21.7}{1.0} & \result{14.6}{0.2} & \result{27.3}{0.4} & \result{25.7}{0.2} \\
 & \multirow{2}{*}{SparseGPT + \ours{}} & Legal & \result{49.1}{0.1} & \result{23.3}{0.1} & \bestresult{29.9}{1.1} & \bestresult{12.7}{0.2} & - & - \\ 
 &  & Biomedical & \bestresult{50.1}{0.1} & \result{22.8}{0.1} & - & - & \bestresult{37.6}{0.5} & \bestresult{23.0}{0.3} \\ \cmidrule{2-9}
 
 & SparseGPT + GPTQ-M & - & \bestresult{47.6}{0.4} & \bestresult{24.5}{0.4} & \result{20.7}{0.7} & \result{16.8}{0.5} & \result{27.7}{1.0} & \result{28.6}{0.2} \\
 & \multirow{2}{*}{SparseGPT + GPTQ-M + \ours{}} & Legal & \result{46.0}{0.6} & \result{25.2}{0.3} & \bestresult{28.9}{2.5} & \bestresult{14.0}{0.3} & - & - \\
 &  & Biomedical & \result{46.1}{0.4} & \result{24.9}{0.3} & - & - & \bestresult{35.6}{0.4} & \bestresult{25.3}{0.4} \\ \midrule

\multirow{7}{*}{Llama 3.1 8B} & - & - & \referenceresult{67.8}{0.0} & \referenceresult{7.3}{0.0} & \referenceresult{55.1}{0.0} & \referenceresult{4.2}{0.0} & \referenceresult{60.8}{0.0} & \referenceresult{8.0}{0.0} \\

 & SparseGPT & - & \result{54.5}{0.2} & \result{17.9}{0.1} & \result{22.1}{1.2} & \result{10.7}{0.2} & \result{34.8}{0.3} & \result{19.3}{0.2} \\
 & \multirow{2}{*}{SparseGPT + \ours{}} & Legal & \result{53.6}{0.1} & \result{18.2}{0.2} & \bestresult{41.5}{0.8} & \bestresult{9.6}{0.1} & - & - \\
 &  & Biomedical & \bestresult{54.9}{0.5} & \bestresult{17.6}{0.1} & - & - & \bestresult{42.4}{0.7} & \bestresult{17.0}{0.2} \\ \cmidrule{2-9}
 
 & SparseGPT + GPTQ-M & - & \bestresult{54.2}{0.3} & \result{19.3}{0.6} & \result{20.7}{1.4} & \result{12.0}{0.2} & \result{33.9}{0.5} & \result{20.9}{0.2} \\
 & \multirow{2}{*}{SparseGPT + GPTQ-M + \ours{}} & Legal & \result{51.9}{0.1} & \result{19.4}{0.2} & \bestresult{39.5}{0.6} & \bestresult{10.8}{0.2} & - & - \\
 &  & Biomedical & \result{52.8}{0.3} & \bestresult{19.0}{0.2} & - & - & \bestresult{39.6}{1.9} & \bestresult{18.2}{0.3} \\ \midrule

\multirow{7}{*}{Gemma 2 2B} & - & - & \referenceresult{63.3}{0.0} & \referenceresult{13.1}{0.0} & \referenceresult{27.1}{0.0} & \referenceresult{6.2}{0.0} & \referenceresult{44.2}{0.0} & \referenceresult{15.0}{0.0} \\

 & SparseGPT & - & \result{47.1}{0.6} & \result{40.0}{1.7} & \result{16.3}{0.0} & \result{17.1}{0.7} & \result{30.4}{0.8} & \result{57.9}{2.4} \\
 & \multirow{2}{*}{SparseGPT + \ours{}} & Legal & \result{48.2}{0.3} & \bestresult{34.4}{0.4} & \bestresult{21.4}{0.5} & \bestresult{13.0}{0.1} & - & - \\
 &  & Biomedical & \bestresult{49.2}{0.4} & \result{36.2}{0.4} & - & - & \bestresult{32.0}{0.8} & \bestresult{40.7}{0.8} \\ \cmidrule{2-9}
 
 & SparseGPT + GPTQ-M & - & \result{45.8}{0.5} & \result{45.2}{2.0} & \result{16.3}{0.1} & \result{19.2}{0.8} & \result{30.4}{2.1} & \result{68.3}{5.8} \\
 & \multirow{2}{*}{SparseGPT + GPTQ-M + \ours{}} & Legal & \result{46.9}{0.2} & \bestresult{37.7}{0.4} & \bestresult{19.4}{0.7} & \bestresult{14.5}{0.2} & - & - \\
 &  & Biomedical & \bestresult{47.6}{0.2} & \result{40.1}{0.5} & - & - & \bestresult{32.5}{0.3} & \bestresult{45.0}{1.7} \\ \midrule

\multirow{7}{*}{Gemma 2 9B} & - & - & \referenceresult{70.2}{0.0} & \referenceresult{10.6}{0.0} & \referenceresult{56.2}{0.0} & \referenceresult{4.8}{0.0} & \referenceresult{62.8}{0.0} & \referenceresult{12.0}{0.0} \\

 & SparseGPT & - & \result{56.9}{1.0} & \result{22.2}{0.5} & \result{32.1}{2.9} & \result{9.6}{0.1} & \result{38.9}{0.9} & \result{30.0}{0.7} \\
 
 & \multirow{2}{*}{SparseGPT + \ours{}} & Legal & \result{57.6}{0.2} & \bestresult{21.1}{0.1} & \bestresult{44.6}{1.1} & \bestresult{7.8}{0.0} & - & - \\
 &  & Biomedical & \bestresult{59.3}{0.4} & \result{21.7}{0.1} & - & - & \bestresult{49.2}{0.6} & \bestresult{23.6}{0.1} \\ \cmidrule{2-9}
 
 & SparseGPT + GPTQ-M & - & \result{56.1}{0.4} & \result{24.3}{0.4} & \result{27.5}{2.8} & \result{10.4}{0.1} & \result{36.5}{1.4} & \result{33.8}{0.6} \\
 & \multirow{2}{*}{SparseGPT + GPTQ-M + \ours{}} & Legal & \result{56.4}{0.5} & \bestresult{22.9}{0.1} & \bestresult{44.2}{1.5} & \bestresult{8.2}{0.1} & - & - \\
 &  & Biomedical & \bestresult{58.5}{0.6} & \result{23.5}{0.1} & - & - & \bestresult{48.9}{1.0} & \bestresult{26.1}{0.2} \\ \midrule

\multirow{7}{*}{Mistral NeMo 12B} & - & - & \referenceresult{69.4}{0.0} & \referenceresult{7.1}{0.0} & \referenceresult{57.5}{0.0} & \referenceresult{4.3}{0.0} & \referenceresult{58.4}{0.0} & \referenceresult{7.6}{0.0} \\

 & SparseGPT & - & \bestresult{57.4}{0.4} & \bestresult{15.6}{0.1} & \result{24.9}{1.4} & \result{9.1}{0.1} & \result{33.1}{1.2} & \result{16.2}{0.1} \\
 & \multirow{2}{*}{SparseGPT + \ours{}} & Legal & \result{56.6}{0.2} & \result{16.3}{0.1} & \bestresult{38.0}{0.5} & \bestresult{8.2}{0.1} & - & - \\
 &  & Biomedical & \bestresult{57.4}{0.1} & \result{16.0}{0.1} & - & - & \bestresult{43.7}{0.3} & \bestresult{15.1}{0.1} \\ \cmidrule{2-9}
 
 & SparseGPT + GPTQ-M & - & \bestresult{56.1}{0.2} & \bestresult{17.3}{0.3} & \result{22.2}{1.5} & \result{10.2}{0.2} & \result{32.0}{2.3} & \result{17.9}{0.3} \\
 & \multirow{2}{*}{SparseGPT + GPTQ-M + \ours{}} & Legal & \result{54.9}{0.3} & \result{18.4}{0.2} & \bestresult{35.2}{2.2} & \bestresult{9.2}{0.2} & - & - \\
 &  & Biomedical & \result{56.0}{0.1} & \result{18.1}{0.5} & - & - &  \bestresult{41.8}{0.6} & \bestresult{16.6}{0.3} \\ \bottomrule

\end{tabular}
    \caption{Average performance for the general and domain-specific benchmarks when pruning with 2:4 sparsity. Standard deviations are shown as subscripts. Rows using the GPTQ-M method additionally employ 4-bit quantization. \textbf{Bold} values denote the best performing method per model for (1) pruning, and (2) joint pruning and quantization. For reference, the top row of each model shows the original model performance (i.e. prior to compression).}
    \label{tab:complete-results-quantization}
\end{table*}

\begin{table*}[t]
    \centering
    \scriptsize
    \begin{tabular}{llll|cc|cc|cc}
\toprule

\multirow{2}{*}{Group Size} & \multirow{2}{*}{Model} & Target & \multirow{2}{*}{$\lambda_g$} & \multicolumn{2}{c|}{General} & \multicolumn{2}{c|}{Legal} & \multicolumn{2}{c}{Biomedical} \\
& & Domain &  & Accuracy & Perplexity & Accuracy & Perplexity & Accuracy & Perplexity \\
\midrule

\multirow{12}{*}[-12pt]{None} & \multirow{6}{*}{Llama 3.2 3B} & Legal &  0.1 & \bestresult{52.4}{0.2} & \result{16.0}{0.0} & \result{27.3}{0.4} & \result{8.2}{0.0} & - & - \\
& & Biomedical &  0.1 & \result{50.9}{0.6} & \result{18.0}{0.0} & - & - & \result{38.7}{1.3} & \result{16.1}{0.1} \\ \cmidrule(lr){3-10}
& & Legal &  0.01 & \result{52.2}{0.4} & \result{16.0}{0.0} & \result{27.3}{0.5} & \result{8.2}{0.0} & - & - \\
& & Biomedical &  0.01 & \result{51.0}{0.6} & \result{17.9}{0.1} & - & - & \bestresult{39.2}{0.9} & \result{16.2}{0.1} \\ \cmidrule(lr){3-10}
& & Legal &  0.001 & \result{52.2}{0.2} & \result{16.0}{0.1} & \bestresult{27.8}{0.6} & \result{8.2}{0.0} & - & - \\
& & Biomedical &  0.001 & \result{50.9}{0.5} & \result{17.8}{0.1} & - & - & \result{38.9}{1.1} & \result{16.1}{0.1} \\  \cmidrule(lr){2-10}

& \multirow{6}{*}{Llama 3.1 8B} & Legal &  0.1 & \result{57.0}{0.3} & \result{12.5}{0.1} & \result{42.2}{2.5} & \result{6.6}{0.0} & - & - \\
& & Biomedical &  0.1 & \result{57.6}{0.7} & \result{14.0}{0.4} & - & - & \result{47.4}{0.3} & \result{15.0}{0.5} \\ \cmidrule(lr){3-10}
& & Legal &  0.01 & \result{57.2}{0.3} & \result{12.6}{0.2} & \bestresult{42.5}{1.8} & \result{6.7}{0.0} & - & - \\
& & Biomedical &  0.01 & \result{58.8}{0.2} & \result{13.3}{0.1} & - & - & \result{48.8}{0.8} & \result{15.1}{0.4} \\ \cmidrule(lr){3-10}
& & Legal &  0.001 & \result{57.0}{0.4} & \result{12.6}{0.2} & \result{41.7}{2.2} & \result{6.7}{0.1} & - & - \\
& & Biomedical &  0.001 & \result{57.7}{1.2} & \result{14.2}{1.0} & - & - & \result{47.5}{1.8} & \result{15.5}{0.5} \\ \midrule

\multirow{12}{*}[-12pt]{128} & \multirow{6}{*}{Llama 3.2 3B}  & Legal & 0.1 & \result{52.0}{0.2} & \result{16.7}{0.1} & \result{27.3}{0.7} & \result{8.2}{0.0} & - & -  \\
 & & Biomedical &  0.1 & \result{52.0}{0.2} & \result{18.9}{0.2} & - & -  & \bestresult{39.2}{0.7} & \result{16.3}{0.1} \\ \cmidrule(lr){3-10}
 & & Legal & 0.01 & \result{52.0}{0.3} & \result{16.6}{0.1} & \result{27.4}{0.6} & \result{8.2}{0.0} & - & - \\
 & & Biomedical &  0.01 & \bestresult{52.1}{0.1} & \result{19.0}{0.2} & - & -  & \result{39.0}{0.6} & \result{16.3}{0.1} \\ \cmidrule(lr){3-10}
 & & Legal & 0.001 & \result{52.1}{0.1} & \result{16.6}{0.1} & \result{27.2}{0.6} & \result{8.2}{0.0} & - & -  \\
 & & Biomedical &  0.001 & \result{52.0}{0.3} & \result{18.9}{0.2} & - & - & \bestresult{39.2}{0.5} & \result{16.2}{0.1} \\ \cmidrule(lr){2-10}
 
& \multirow{6}{*}{Llama 3.1 8B}  & Legal & 0.1 & \result{57.3}{0.3} & \result{12.9}{0.1} & \result{42.0}{2.0} & \result{6.6}{0.0} & - & -  \\
&  & Biomedical &  0.1 & \result{58.1}{0.8} & \result{14.2}{0.4} & - & - & \result{47.8}{0.8} & \result{14.7}{0.5} \\ \cmidrule(lr){3-10}
&  & Legal & 0.01 & \bestresult{57.4}{0.3} & \result{12.9}{0.2} & \result{42.1}{1.3} & \result{6.6}{0.0} & - & - \\
&  & Biomedical &  0.01 & \bestresult{59.1}{0.2} & \result{13.5}{0.1} & - & -  & \bestresult{49.2}{0.4} & \result{14.7}{0.3} \\ \cmidrule(lr){3-10}
&  & Legal & 0.001 & \result{57.1}{0.3} & \result{13.0}{0.2} & \result{40.6}{2.6} & \result{6.7}{0.1} & - & -  \\
&  & Biomedical &  0.001 & \result{58.1}{1.2} & \result{14.3}{0.9} & - & - & \result{47.8}{1.7} & \result{15.1}{0.6} \\ 
\bottomrule
\end{tabular}
    \caption{Average D-Pruner performance for general and domain-specific benchmarks. We vary the group size (i.e. iterative blocking) and regularization ($\lambda_g$) hyperparameters. Standard deviations are shown as subscripts. \textbf{Bold} values denote the hyperparameter combination with the highest accuracy for each model and target domain.}
    \label{tab:complete-results-dpruner}
\end{table*}

\begin{table*}[t]
    \centering
    \scriptsize
    \begin{tabular}{ll|cccc|cccc}
\toprule

\multirow{2}{*}{Model} & \multirow{2}{*}{Method} & \multicolumn{4}{c|}{Biomedical} & \multicolumn{3}{c}{Legal}\\

& & MedMCQA & MedQA (4) & PubMedQA & Bio. MMLU & CaseHOLD & ECtHR & Legal MMLU \\
\midrule

\multirow{6}{*}{Llama 3.2 3B} & -  & \referenceresult{49.5}{0.0} & \referenceresult{51.5}{0.0} & \referenceresult{72.8}{0.0} & \referenceresult{61.1}{0.0} & \referenceresult{42.7}{0.0} & \referenceresult{49.6}{0.0} & \referenceresult{44.2}{0.0} \\
& Magnitude   & \result{28.8}{0.0} & \result{28.8}{0.0} & \result{50.6}{0.0} & \result{27.5}{0.0} & \result{20.6}{0.0} & \result{4.5}{0.0} & \result{27.2}{0.0} \\
& Wanda & \result{30.4}{0.2} & \result{35.6}{0.6} & \result{63.6}{0.5} & \result{41.8}{0.3} & \result{28.9}{0.7} & \result{13.7}{0.7} & \result{35.4}{0.4} \\
& SparseGPT  & \result{33.6}{1.9} & \result{35.4}{1.4} & \result{69.5}{1.2} & \result{45.1}{1.2} & \result{32.9}{0.6} & \result{15.8}{1.6} & \result{37.1}{0.5} \\
& D-Pruner  & \result{35.3}{0.8} & \result{39.2}{0.3} & \result{66.8}{1.2} & \result{41.5}{0.6} & \result{22.9}{1.3} & \result{25.6}{0.1} & \result{33.1}{0.7} \\
 & SparseGPT + \ours{} & \bestresult{41.1}{0.4} & \bestresult{41.9}{0.8} & \bestresult{70.8}{0.8} & \bestresult{51.3}{1.1} & \bestresult{40.2}{1.3} & \bestresult{28.3}{3.5} & \bestresult{40.8}{0.8} \\ \midrule

\multirow{6}{*}{Llama 3.1 8B} & - & \referenceresult{56.4}{0.0} & \referenceresult{60.1}{0.0} & \referenceresult{75.8}{0.0} & \referenceresult{71.7}{0.0} & \referenceresult{51.9}{0.0} & \referenceresult{60.9}{0.0} & \referenceresult{52.4}{0.0} \\
 & Magnitude & \result{32.4}{0.0} & \result{34.2}{0.0} & \result{59.6}{0.0} & \result{37.0}{0.0} & \result{20.4}{0.0} & \result{10.0}{0.0} & \result{32.2}{0.0} \\
 & Wanda & \result{38.5}{0.3} & \result{38.0}{0.9} & \result{66.8}{0.5} & \result{51.3}{0.6} & \result{29.4}{0.7} & \result{35.9}{0.3} & \result{39.7}{0.3} \\
 & SparseGPT & \result{41.0}{1.8} & \result{43.2}{1.0} & \result{70.7}{1.0} & \result{55.7}{1.1} & \result{38.5}{1.5} & \result{38.0}{1.5} & \result{43.9}{0.5} \\
 & D-Pruner & \result{43.4}{1.5} & \result{45.1}{2.1} & \result{69.9}{2.8} & \result{57.5}{2.0} & \result{37.1}{2.4} & \result{41.4}{6.6} & \result{43.4}{0.8} \\
 & SparseGPT + \ours{} & \bestresult{48.3}{0.5} & \bestresult{49.2}{0.8} & \bestresult{72.3}{0.9} & \bestresult{63.4}{0.6} & \bestresult{45.6}{0.5} & \bestresult{54.2}{1.4} & \bestresult{46.6}{1.0} \\ \midrule
 
\multirow{5}{*}{Gemma 2 2B} &  - & \referenceresult{40.9}{0.0} & \referenceresult{35.3}{0.0} & \referenceresult{74.0}{0.0} & \referenceresult{53.8}{0.0} & \referenceresult{32.5}{0.0} & \referenceresult{8.5}{0.0} & \referenceresult{40.4}{0.0} \\
& Magnitude& \result{22.5}{0.0} & \result{24.1}{0.0} & \result{56.4}{0.0} & \result{32.9}{0.0} & \result{21.2}{0.0} & \bestresult{17.9}{0.0} & \result{25.9}{0.0} \\
& Wanda & \result{25.8}{0.9} & \result{24.4}{0.9} & \result{57.4}{0.6} & \result{37.6}{1.0} & \result{19.8}{0.1} & \result{12.2}{4.8} & \result{28.3}{0.6} \\
& SparseGPT & \result{26.0}{2.1} & \result{27.7}{2.0} & \result{60.4}{1.9} & \result{36.9}{1.2} & \result{24.0}{2.1} & \result{6.2}{2.4} & \result{32.3}{1.7} \\
& SparseGPT + \ours{} & \bestresult{36.6}{0.2} & \bestresult{33.4}{1.1} & \bestresult{67.0}{1.1} & \bestresult{45.8}{0.6} & \bestresult{32.2}{2.2} & \result{5.1}{1.1} & \bestresult{37.8}{0.3} \\ \midrule

\multirow{5}{*}{Gemma 2 9B}& - & \referenceresult{57.9}{0.0} & \referenceresult{60.5}{0.0} & \referenceresult{78.6}{0.0} & \referenceresult{77.2}{0.0} & \referenceresult{51.7}{0.0} & \referenceresult{60.1}{0.0} & \referenceresult{56.8}{0.0} \\
& Magnitude & \result{43.7}{0.0} & \result{46.3}{0.0} & \result{71.8}{0.0} & \result{54.5}{0.0} & \result{37.1}{0.0} & \result{19.9}{0.0} & \result{44.1}{0.0} \\
& Wanda  & \result{45.0}{0.5} & \result{44.7}{1.5} & \result{71.4}{0.7} & \result{57.7}{0.7} & \result{43.6}{1.6} & \result{36.1}{2.1} & \result{45.8}{1.0} \\
& SparseGPT & \result{48.0}{0.5} & \result{48.6}{0.9} & \result{74.4}{1.1} & \result{66.1}{0.4} & \result{45.8}{1.5} & \result{36.2}{0.9} & \result{49.2}{0.9} \\
& SparseGPT + \ours{} & \bestresult{53.0}{0.4} & \bestresult{54.1}{1.1} & \bestresult{76.4}{0.9} & \bestresult{70.5}{0.2} & \bestresult{49.8}{0.1} & \bestresult{55.0}{0.4} & \bestresult{53.6}{0.5} \\ \midrule

\multirow{5}{*}{Mistral NeMo 12B} & - & \referenceresult{52.4}{0.0} & \referenceresult{60.6}{0.0} & \referenceresult{74.4}{0.0} & \referenceresult{71.9}{0.0} & \referenceresult{55.8}{0.0} & \referenceresult{62.0}{0.0} & \referenceresult{54.8}{0.0} \\
 & Magnitude  & \result{30.0}{0.0} & \result{33.0}{0.0} & \result{67.0}{0.0} & \result{42.3}{0.0} & \result{25.8}{0.0} & \result{9.8}{0.0} & \result{37.9}{0.0} \\
 & Wanda  & \result{39.7}{0.4} & \result{45.5}{0.3} & \result{60.6}{0.1} & \result{59.7}{0.4} & \result{44.0}{0.5} & \result{37.4}{2.8} & \result{45.4}{0.5} \\
 & SparseGPT& \result{40.0}{0.5} & \result{46.5}{0.9} & \result{72.0}{0.7} & \result{59.2}{1.5} & \result{40.5}{3.1} & \result{49.3}{3.4} & \result{43.8}{0.9} \\
 & SparseGPT + \ours{} & \bestresult{47.7}{0.7} & \bestresult{55.3}{0.4} & \bestresult{73.2}{0.5} & \bestresult{65.2}{0.7} & \bestresult{48.3}{2.8} & \bestresult{55.1}{1.1} & \bestresult{49.9}{0.5} \\

\bottomrule
\end{tabular}
    \caption{Average performance across calibration sets for tasks from the biomedical and legal domains. Standard deviations are shown as subscripts. \textbf{Bold} values denote the best performing method for each task.}
    \label{tab:complete-results-analytical}
\end{table*}

\begin{table*}[t]
    \centering
    \scriptsize
    \begin{tabular}{lll|ccc}
\toprule

Model & Method & Target Domain & General & Legal & Biomedical \\
\midrule

\multirow{6}{*}{Llama 3.2 3B}

& SparseGPT & -  & \result{13.8}{0.1} & \result{8.2}{0.1} & \result{14.7}{0.1} \\ \cmidrule{2-6}
& \multirow{2}{*}{+ In-domain data} & Legal & \result{13.7}{0.0} & \bestresult{7.6}{0.1} & - \\
&  & Biomedical & \result{13.5}{0.0} & - & \result{13.7}{0.1} \\  \cmidrule{2-6}
& \multirow{2}{*}{+ \ours{} (Ours)} & Legal & \bestresult{13.6}{0.0} & \result{7.7}{0.2} & - \\
&  & Biomedical & \bestresult{13.3}{0.0} & - & \bestresult{13.5}{0.1} \\ \midrule

\multirow{6}{*}{Llama 3.1 8B}
 
& SparseGPT  & - & \bestresult{10.7}{0.1} & \result{6.3}{0.1} & \result{11.5}{0.1} \\ \cmidrule{2-6}
& \multirow{2}{*}{+ In-domain data} & Legal & \result{10.8}{0.0} & \bestresult{6.0}{0.0} & - \\
&  & Biomedical & \result{10.7}{0.0} & - & \result{10.9}{0.0} \\  \cmidrule{2-6}
& \multirow{2}{*}{+ \ours{} (Ours)} & Legal & \bestresult{10.7}{0.0} & \bestresult{6.0}{0.1} & - \\
&  & Biomedical & \bestresult{10.5}{0.0} & - & \bestresult{10.7}{0.0} \\ \midrule

\multirow{6}{*}{Gemma 2 2B}
& SparseGPT & - & \result{20.8}{2.0} & \result{9.3}{0.8} & \result{24.3}{1.8} \\ \cmidrule{2-6}
& \multirow{2}{*}{+ In-domain data}  & Legal & \result{19.9}{0.1} & \bestresult{8.3}{0.0} & - \\
&  & Biomedical & \result{20.5}{0.1} & - & \result{23.5}{0.2} \\ \cmidrule{2-6}
& \multirow{2}{*}{+ \ours{} (Ours)} & Legal & \bestresult{18.9}{0.1} & \bestresult{8.3}{0.0} & - \\
&  & Biomedical & \bestresult{19.1}{0.1} & - & \bestresult{21.5}{0.2} \\ \midrule

\multirow{6}{*}{Gemma 2 9B} 
& SparseGPT & - & \result{15.2}{0.2} & \result{6.4}{0.0} & \result{19.3}{0.2} \\ \cmidrule{2-6}
& \multirow{2}{*}{+ In-domain data} & Legal & \result{14.7}{0.0} & \bestresult{5.9}{0.0} & - \\
& & Biomedical & \result{14.9}{0.0} & - & \result{17.5}{0.1} \\ \cmidrule{2-6}
& \multirow{2}{*}{+ \ours{} (Ours)} & Legal & \bestresult{14.6}{0.0} & \bestresult{5.9}{0.0} & - \\
&  & Biomedical & \bestresult{14.5}{0.1} & - & \bestresult{16.8}{0.1} \\ \midrule

\multirow{6}{*}{Mistral NeMo 12B} 
& SparseGPT & - & \bestresult{9.4}{0.0} & \result{5.6}{0.0} & \result{9.8}{0.0} \\ \cmidrule{2-6}
& \multirow{2}{*}{+ In-domain data}  & Legal & \result{9.5}{0.0} & \bestresult{5.5}{0.0} & - \\
&  & Biomedical & \result{9.5}{0.0} & - & \result{9.7}{0.0} \\ \cmidrule{2-6}
& \multirow{2}{*}{+ \ours{} (Ours)} & Legal & \result{9.5}{0.0} & \bestresult{5.5}{0.0} & - \\
&  & Biomedical & \bestresult{9.4}{0.0} & - & \bestresult{9.6}{0.0} \\

\bottomrule
\end{tabular}
    \caption{Average performance when pruning to 50\% unstructured sparsity with (1) in-domain data, and (2) \ours{}. Standard deviations are shown as subscripts. \textbf{Bold} values denote the best performing method for each model.}
    \label{tab:complete-results-ablation}
\end{table*}

\end{document}